\documentclass[10pt,twocolumn,letterpaper]{article}

\usepackage{cvpr}              

\usepackage{graphicx}
\usepackage{amsmath}
\usepackage{amssymb}
\usepackage{booktabs}
\usepackage{multirow}

\usepackage[ruled, vlined]{algorithm2e}

\newtheorem{definition}{Definition}

\newtheorem{theorem}{Theorem}

\usepackage[pagebackref,breaklinks,colorlinks,bookmarks=false]{hyperref}

\usepackage[capitalize]{cleveref}
\crefname{section}{Sec.}{Secs.}
\Crefname{section}{Section}{Sections}
\Crefname{table}{Table}{Tables}
\crefname{table}{Tab.}{Tabs.}

\def\cvprPaperID{*****} 
\def\confName{CVPR}
\def\confYear{2022}

\begin{document}

\title{Federated Learning with Label Distribution Skew via Logits Calibration}

\author{Jie Zhang$^{1}$\quad Zhiqi Li$^1$ \quad Bo Li$^{2\ddagger}$\quad Jianghe Xu$^{2}$\quad Shuang Wu $^{2}$\quad \\ Shouhong Ding$^{2}$\quad Chao Wu$^{1\ddagger}$\\
\\
$^1$Zhejiang University \qquad $^2$Youtu Lab, Tencent \\
}

\maketitle

\begin{abstract}
Traditional federated optimization methods perform poorly with heterogeneous data (i.e.\ , accuracy reduction), especially for highly skewed data.  In this paper, we investigate the label distribution skew in FL, where the distribution of labels varies across clients. First, we investigate the label distribution skew from a statistical view. We demonstrate both theoretically and empirically that previous methods based on softmax cross-entropy are not suitable, which can result in local models heavily overfitting to minority classes and missing classes. Additionally, we theoretically introduce a deviation bound to measure the deviation of  the gradient after local update. At last, we propose FedLC (\textbf{Fed}erated learning via \textbf{L}ogits \textbf{C}alibration), which calibrates the logits  before softmax cross-entropy according to the probability of occurrence of each class. FedLC applies a fine-grained calibrated cross-entropy loss into local update by adding a pairwise label margin. Extensive experiments on federated datasets and real-world datasets demonstrate that FedLC leads to a more accurate global model and much improved performance. Furthermore, integrating other FL methods into our approach can further enhance the performance of the global model.
\end{abstract}

\section{Introduction}
Recently, machine learning techniques enriched by massive data have been used as a fundamental technology with applications in both established and emerging fields~\cite{DBLP:conf/ijcai/0061STSS19,DBLP:conf/aaai/LiSG19,DBLP:conf/iccv/LiSLWH19,DBLP:conf/mm/LiSWL19,DBLP:conf/icassp/LiSTH19,DBLP:conf/accv/Tang020,DBLP:journals/corr/abs-2104-14729,DBLP:conf/mmm/HuSLYL17,DBLP:conf/iccv/TangLZDS21,DBLP:journals/corr/abs-2110-12748,DBLP:conf/mm/LiXWDLH21,tang2022re,Zhong_2022_CVPR,Chen_2022_CVPR,Zhang_2022_CVPR}. 
A large volume of data is generated on various edge devices, raising concerns about privacy and security. Federated learning (FL)~\cite{mcmahan2017communication} emerges as a new distributed learning paradigm, which is concerned with privacy data that distributed in a non-IID (not Independent and Identically Distributed) manner. 

\begin{figure}[t] 
    \centering
    \includegraphics[width=8cm]{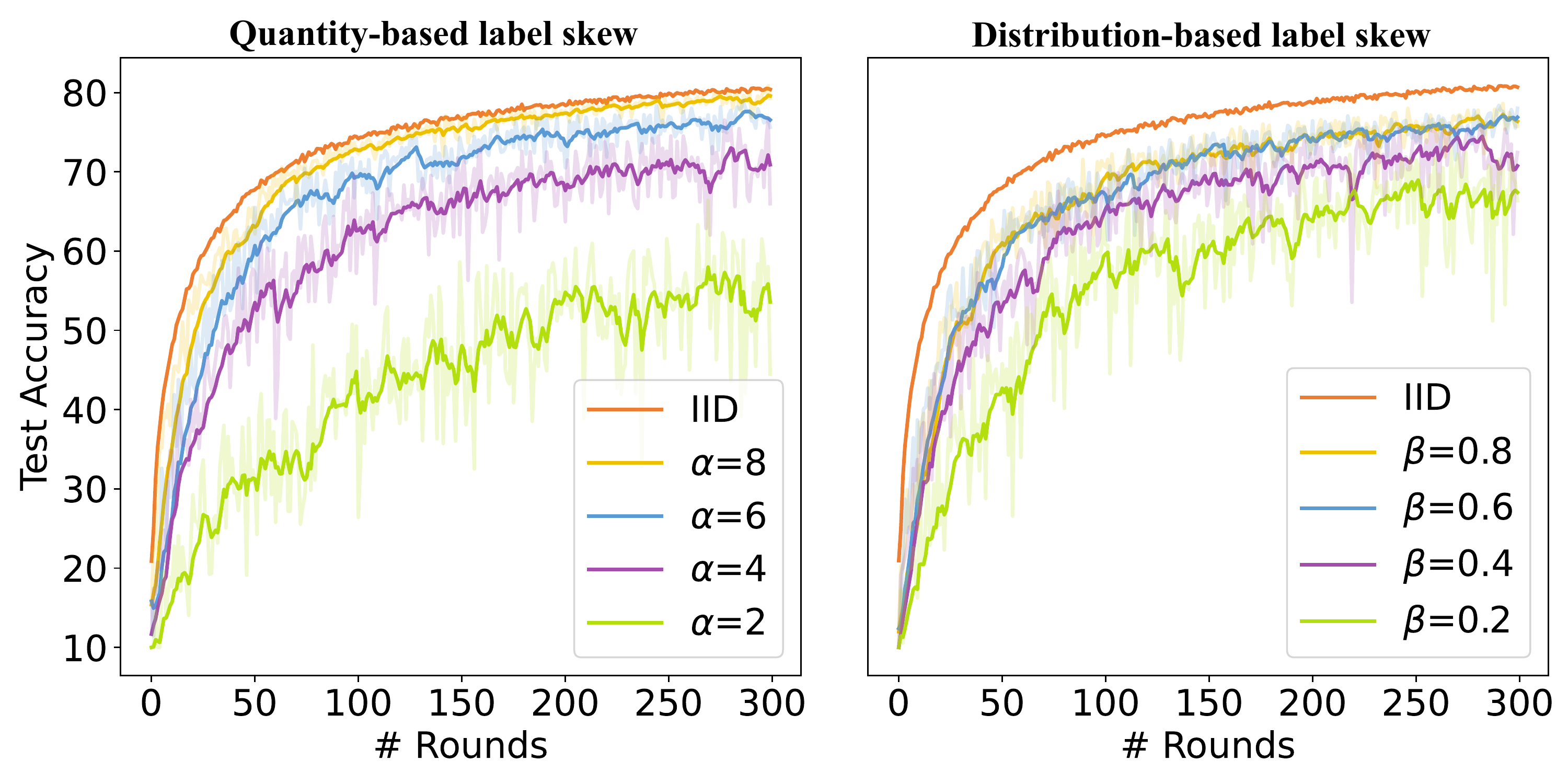}
    \vspace{-0.4cm}
    \caption{Test accuracy of FedAvg under various label skew settings on CIFAR10. The lower the $\alpha$ and $\beta$, the more skewed the distribution. In comparison with IID settings, the accuracy is significantly decreased by 26.07 \% and 13.97 \% for $\alpha=2$ and $\beta=0.2$, respectively.}
    \label{intro_fig}
    \vspace{-0.4cm}
\end{figure}

As mentioned in earlier studies~\cite{zhao2018federated,kairouz2021advances,li2018federated}, heterogeneous data can degrade the effectiveness of FL.   Recent studies have proposed many methods to solve the issue of accuracy degradation in non-IID settings, such as FedProx~\cite{li2018federated}, Scaffold~\cite{karimireddy2020scaffold}, FedNova~\cite{DBLP:conf/nips/WangLLJP20} and FedOpt~\cite{DBLP:conf/iclr/ReddiCZGRKKM21}.  However, previous studies have very rigid data partitioning strategies among 
clients, which are hardly representative and thorough. To better explore the effect of non-IID data, ~\cite{li2021federated} develops a benchmark with more comprehensive non-IID settings (e.g.\ , label distribution skew, feature distribution skew). As demonstrated in~\cite{li2021federated}, none of
these existing state-of-the-art FL algorithms outperforms others in all non-IID settings. It inspires researchers to develop specialized algorithms for a specific non-IID setting to further improve the performance of the global model. For example, FedBN~\cite{li2021fedbn} aims to address the feature distribution skew in FL.

In this paper, we primarily investigate the \textbf{label distribution skew}\footnote{To simulate label distribution skew, we conduct comprehensive experiments with different degrees of non-iidness.} in FL. Label distribution skew is one of the most challenging non-IID settings, where the distribution of labels varies across clients~\cite{kairouz2021advances, li2021federated}. In fact, label distribution skew always exists in real-world applications~\cite{wang2021field,alshedivat2021federated}. For example, pandas are only found in China and zoos, and a person's face may only appear in a few places worldwide.  Following the settings in previous works ~\cite{li2021federated,kairouz2021advances}, we simulate two different label skew scenarios that are commonly used in practice: quantity-based label skew and distribution-based label skew (see detailed description in Section~\ref{statement}). As shown in Figure~\ref{intro_fig}, in comparison with IID setting, the test accuracy is significantly decreased by 26.07 \% and 13.97 \% for highly skewed data~\footnote{The more skewed the data distribution, the more difficult it is to improve the performance of the global model.}. We argue that this is reasonable. As demonstrated in previous studies~\cite{DBLP:conf/nips/WangLLJP20}, heterogeneous data can result in 
inconsistent objective functions among clients, which leads the global model to converge to a stationary point that is far from global optima~\cite{DBLP:conf/nips/WangLLJP20}. 
Furthermore, skewed data on the local client results in a biased model overfitting to minority classes and missing classes, which aggravates the objective inconsistency between clients (see the discussion in Section~\ref{sec3}). Therefore, aggregating these severely biased models can result in the global model being further away from the optimal solution.

Previous studies have attempted to solve the inter-client objective inconsistency by regularizing the local objectives~\cite{li2018federated, li2020feddane, acar2021federated, zhang2020fedpd}, while ignoring the impact of local skewed data. Instead, our intuitive solution is to address the negative effects caused by intra-client label skew, which aims to reduce the bias in local update
and in turn benefits the global model.
This is because the performance of the global model is highly dependent on local models. Thus, resolving the intra-client label skew will produce higher quality of local models, and then a greater performance for the global model. 


Our contributions are summarized as follows: 1) we first investigate the label distribution skew from a statistical perspective, and demonstrate that previous methods based on softmax cross-entropy are not suitable, which can result in biased local models. 2) Then we theoretically introduce a deviation bound to measure the deviation  of  the gradient after local update. 3) At last, we propose FedLC (\textbf{Fed}erated learning via \textbf{L}ogits \textbf{C}alibration), which calibrates the logit of each class before softmax cross-entropy according to the probability of occurrence. 
In detail, FedLC applies a fine-grained calibrated cross-entropy loss into local update by adding a pairwise label margin.
By forcing the training to focus on the margins of missing classes and minority classes to reach the optimal threshold, our method encourages these underrepresented classes  to have larger margins. 
4) We show both theoretically and empirically that FedLC leads to a more accurate global model and much improved performance. Furthermore, integrating other methods that address inter-client objective inconsistency with our approach can further improve the performance of the server model.

\section{Related Works}
\paragraph{Federated Learning with Non-IID Data}  In FL,  the non-IID property across heterogeneous clients makes the local update diverge a lot, posing a fundamental challenge to aggregation. The performance of federated learning suffers from the heterogeneous data located over multiple clients~\cite{acar2021federated, li2020feddane}.~\cite{zhao2018federated} demonstrates that the accuracy of federated learning reduces significantly when models are trained with highly skewed data, which is explained by diverging weights.
In a similar way to FedAvg, FedProx~\cite{li2018federated} utilizes partial information aggregation and proximal term to deal with heterogeneity. FedNova~\cite{DBLP:conf/nips/WangLLJP20} puts insight on the number of epochs in local updates and proposes a normalized averaging scheme to eliminate objective inconsistency. FedOpt~\cite{DBLP:conf/iclr/ReddiCZGRKKM21} proposes to uses adaptive server optimization in FL and Scaffold~\cite{karimireddy2020scaffold} uses control variates (variance reduction) to correct for the client-drift in its local updates. 
However, these previous works treat the non-IID problem as a general problem. How to design an effective FL algorithm to mitigate the significant accuracy reduction for label distribution skew still remains largely unexplored.

\paragraph{Learn from Imbalanced data}
In recent years, many studies have been focusing on analyzing imbalanced data~\cite{he2009learning,liu2019large,zhang2021bag}. Real-world data usually exhibits a imbalanced distribution, and the effectiveness  of machine learning is severely affected by highly skewed data~\cite{jamal2020rethinking,cao2019learning}. Re-sampling~\cite{chawla2002smote} and re-weighting~\cite{kim2020adjusting,jamal2020rethinking,cui2019class} are traditional methods for addressing imbalanced data.  Recent  works use re-weighting methods to enable networks to pay more attention to minority categories by assigning a variable weight to each class. Besides, over-sampling minority classes and under-sampling frequent classes are two re-sampling methods that have been extensively discussed in previous studies. New perspectives like decoupled training~\cite{kang2019decoupling} and deferred re-balancing~\cite{cao2019learning} schedule are also proved to be effective. The majority of previous works on imbalanced data focus on long-tailed distributions. However, in federated learning settings, data can be imbalanced in many ways~\cite{kairouz2021advances,li2021federated}, such as quantity-based label imbalance and distribution-based label imbalance, as discussed in this paper. Besides, the label distribution skew includes long-tail scenarios, but long-tailed methods cannot handle the issue of missing classes, which is extremely common in FL.

\paragraph{Federated Learning with Label Distribution Skew} 
To alleviate the negative effect of label distribution skew, ~\cite{wang2021addressing} proposes a monitoring scheme to detect class imbalance in FL. Nevertheless, this method relies heavily on auxiliary data, which is not practical in real-world FL and poses potential privacy issues. Besides, this method also needs an additional monitor in the central server, which requires more computation. 
Note that FedRS~\cite{li2021fedrs} works in a similar manner, which also attempts to alleviate the negative effect caused by local training on label skewed data. However, it only resolves the issue of missing classes during local updates. Generally, in real-world applications, the local data contains both majority classes and minority classes, as well as missing classes. By contrast with previous methods, our approach systematically analyzes the problem of federated learning with label distribution skew from a statistical perspective. Our approach considers majority classes, minority classes, and missing classes at the same time, which is more practical.
\begin{figure*}[t] 
    \centering
    \includegraphics[width=17cm]{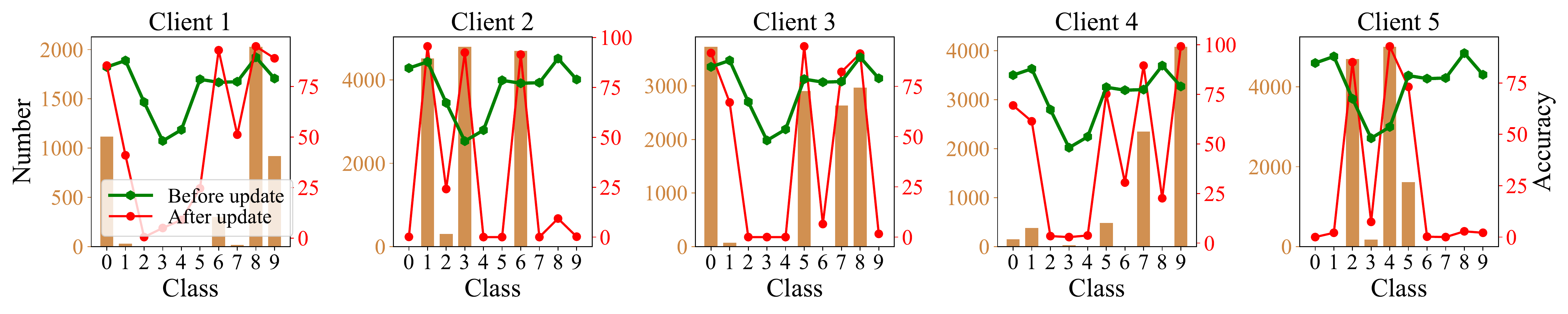}
     \vspace{-0.5cm}
    \caption{For skewed CIFAR10 dataset, the accuracy decreases heavily on minority classes, achieving an overall accuracy of zero for missing classes. The histogram displays the number of samples for each class, while the red line represents the accuracy of each class. }
    \label{overfit_fig}
\end{figure*}
\begin{figure*}[t]
    \centering
    \includegraphics[width=16cm]{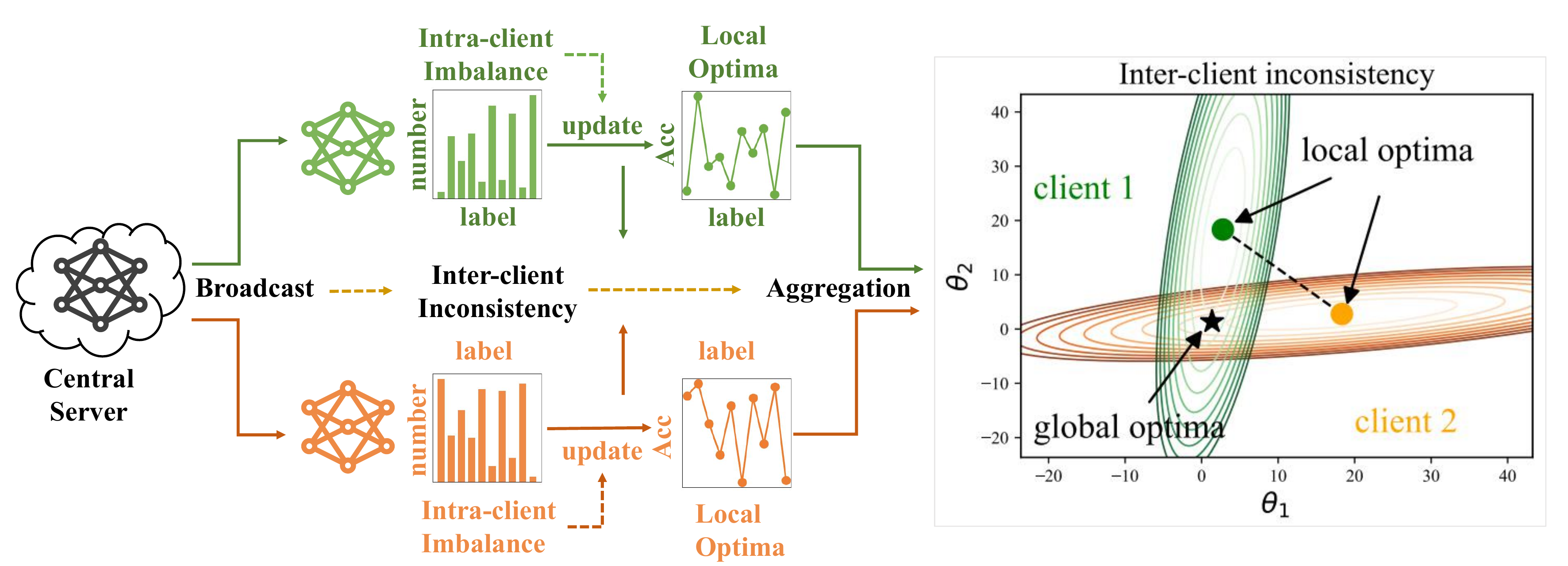}
    \caption{A toy example with 2 clients. The intra-client skew results in a biased model (converge to local optima), which means that after local update the accuracy drops heavily on minority classes.
    Moreover, averaging such biased local models can lead to a poor model that stray from the corresponding global optima~ \protect\cite{alshedivat2021federated}.}
    \label{reason_for_acc_drop}  
\end{figure*}

\section{FL with Label Distribution Skew}
\label{sec3}
\begin{definition}[\textbf{Label Distribution Skew }]
Suppose that client $i$ can draw an example $(x, y) \sim P_{i}(x, y)$ from the local data, and the data distribution ${P}_{i}(x, y)$ can be rewritten as $P_{i}(x \mid y)P_{i}(y)$. For label distribution skew, the marginal distributions $P_i(y)$  varies across clients, while $P_{i}(y \mid x)=P_{j}(y \mid x)$ for all clients $i$ and $j$.

\end{definition}
\begin{definition}[\textbf{Majority, Minority and Missing Classes}]
Generally, for a label skewed dataset, the label set $K$ is split into majority class $j\in \mathcal{J}$, minority class $r\in \mathcal{R}$ and missing class $s\in \mathcal{S}$ respectively. We have $\mathcal{J} \cap \mathcal{R}=\emptyset, \mathcal{R} \cap \mathcal{S} =\emptyset, \mathcal{S}\cap\mathcal{J}=\emptyset   \text { and } \mathcal{J} \cup \mathcal{R} \cup \mathcal{S}=K$, where $\emptyset$ is an empty set. The number of training samples in each class satisfies $n_{j}\gg n_{r}>0, n_{s}=0$.

 
\end{definition}

In federated learning, a total of $m$ clients
aim to jointly minimize the following optimization problem:
$$
\min _{\boldsymbol{x} \in \mathbb{R}^{d}} F(\boldsymbol{x}):=\sum_{i=1}^{m} p_{i} F_{i}(\boldsymbol{x}), F_{i}(\boldsymbol{x})=\frac{1}{n_{i}} \sum_{\xi \in \mathcal{P}_{i}} f_{i}(\boldsymbol{x} ; \xi), 
$$
where $F_{i}(\boldsymbol{x})$ is the objective function of $i$-th client, and $p_{i}$ denotes the relative sample size and $\sum_{i=1}^{m} p_{i}=1$. The local data distribution $P_i$ varies among clients, posing the problem of data heterogeneity.  In FL, the clients selected in each communication round perform multiple local updates, and then these models are aggregated to update a global model. However, for label skew settings, after local updates, these models will learn highly biased decision boundaries, resulting in poor performance when aggregated. 

Affected by the heterogeneous data, the local objectives at clients are generally not identical and may not share same minimizers~\cite{wang2021field}. Thus, when updating local models from the same global model, clients will drift towards the minima of local objectives. This phenomenon is often referred to as client drift~\cite{DBLP:conf/aistats/CharlesK21, karimireddy2020scaffold}. As demonstrated in previous studies~\cite{DBLP:conf/nips/WangLLJP20,zhao2018federated}, standard averaging of these models with client drift leads to convergence to a stationary point which is not that of the original objective function, but of an inconsistent objective function. That is, federated learning cannot achieve optimal weights when training data distribution is skewed. Generally, the more skewed the local data, the harder it is to aggregate a well-performed global model.





Besides, we empirically verify the impact of label skew in FL on five clients after 400 communication rounds and test the accuracy for each class before and after local update. All local models on the test set have same test accuracy before local update (currently, these local models are equivalent to the global model). As illustrated in Figure~\ref{overfit_fig}, after training on local data, the test accuracy of majority classes is even higher than that of the global model, and the test accuracy of minority classes is much lower, and the accuracy of missing classes is almost zero. It indicates that label skew can lead to a biased model, which severely overfits to the minority classes and missing classes. As a result, label skew exacerbates the negative effects of objective inconsistency and leads to a more inaccurate global model.


Based on above discussions, we show an illustration in Figure~\ref{reason_for_acc_drop} of why FL with label distribution skew performs poorly. In the following section, we provide a deeper understanding of label skewed data from a statistic view. 
\paragraph{Deviation of Standard FL algorithms} 
To better analyse the negative effect of label distribution skew in FL, we theoretically show the problems of local update when faced with label distribution skew. We focus on multi-classification task for each client with label set $K$. Let $f(x)=\{w_{y}^{T} h\}_{y \in K}$ be the score function for a given input $x$, $w$ is the classification weights in the last layer and $h$ is the extracted feature as the input of the last layer. Let $p=\sigma(f(x))$ denote the probability vector after softmax function $\sigma(\cdot)$. To evaluate the degree of deviation of the update $\{\Delta w_{y}\}_{y \in K}$ from the expected direction during the local training, we introduce the deviation bound as follows:
\begin{definition}[\textbf{Deviation Bound}] \label{deviationBound}
For majority class $j \in \mathcal{J}$ and minority class $r \in \mathcal{R}$, let  $O_j$ and $O_r$  denote the set of training samples belong to class $j$ and class $r$, respectively. The deviation bound is :
$$D_{jr}=\frac{(1-\overline{p_r^{(r)}})||\overline{h^{(r)}}||_2^2}{\overline{p^{(j)}_{r}}\overline{h^{(r)}}\cdot\overline{h^{(j)}}}=\sum_{\substack{y \in K, y \neq r}   }\frac{\overline{p_{y}^{(r)}}||\overline{h^{(r)}}||_2^2}{\overline{p^{(j)}_{r}}\overline{h^{(r)}}\cdot\overline{h^{(j)}}}, $$ where $\overline{p_{r}^{(j)}}=\frac{1}{n_j}\sum_{i\in O_j} p_{r}(x_{i}),$  $\overline{h^{(j)}}=\frac{1}{n_j}\sum_{i\in O_j} h(x_{i})$ are averaged over all samples in majority class $j$, $\overline{p_{r}^{(r)}}=\frac{1}{n_r}\sum_{i\in O_r} p_{r}(x_{i})$, $\overline{p_{y}^{(r)}}=\frac{1}{n_r}\sum_{i\in O_r} p_{y}(x_{i})$ and $\overline{h^{(r)}}=\frac{1}{n_r}\sum_{i\in O_r} h(x_{i})$ are averaged over all samples in minority  class $r$. 
\end{definition}


\begin{theorem} \label{firstTheorem}
For majority class $j \in \mathcal{J}$ and minority class $r \in \mathcal{R}$.  When $n_{j}/n_r \gg D_{jr} > 0$, the update $\{\Delta w_y\}_{y\in K}$ is much more likely to deviate from expected direction, where ${\Delta w_r} \overline{h^{(r)}}<0$ and $\Delta w_k \overline{h^{(r)}}>0$. 
\end{theorem}
Here, the relation between $n_{j}/n_{r}$ and $D_{jr}$ describes how the scores for majority classes overwhelm those for minority classes during local training, and it generalise previous work~\cite{li2021fedrs}, which only considers missing classes. Missing class can be viewed as a special case of minority class, where $n_r \approx 0$, and $n_{j}/n_r \gg D_{jr}$ almost always holds.

Based on above discussions, the deviation bound gives us a quantitative perspective to reveal the drawbacks of previous FL methods based on standard  FL (softmax cross-entropy) during the local update, where the update $\{\Delta w_y\}_{y\in K}$ deviates from expected direction.  

\section{Federated Learning via Logits Calibration}
As mentioned before, the local update is biased with label distribution skew. In this section, we demonstrate that standard softmax cross-entropy is not suitable for local update with highly skewed data.  To overcome this, we propose a fine-grained calibrated cross-entropy loss to reduce the bias in local update.

\paragraph{Learning Objective}  
Suppose the data distribution at $i$-th client is ${P}_{i}(x, y)=P_{i}(x \mid y) P_{i}(y)$.  Given a data $x$, the predicted label is $\hat{y}=\arg \max_y f_y(x)$.  For balanced label distribution, the goal of standard machine learning is to minimize the misclassification error from a statistical perspective:
\begin{equation}
{P}_{x,y}(y\neq \hat{y}), \text{and  } {P}(y\mid x) \propto P(x \mid y)  P(y).
\end{equation}
Since softmax cross-entropy is usually chosen as the surrogate loss function, the probability $P_y(x)\propto e^{f_y(x)}$ is regarded as the estimates of ${P}(y\mid x)$. 

However, in this paper, we focus on label distribution skew in FL, which means ${P}(y)$ is skewed. That is, minority classes have a much lower probability of occurrence compared with majority classes, which means that minimizing the misclassification error $P(x \mid y)  P(y)$ is no longer suitable~\cite{menon2021longtail}. To handle this, we average each of the per-class error rate~\cite{menon2021longtail}, and attempt to minimize the test error as follows:
\begin{equation}
\text{Calibrated error} =\min \frac{1}{k}\sum_{y\in K}\mathcal{P}_{x \mid y}(y\neq\hat{y}).
\end{equation}
In this manner, the result is the estimate of $P(x\mid y)$, thus 
varying $P(y)$ arbitrarily will not affect the optimal results. 
In other words, when label distribution is skewed, we aim to minimize the calibrated error ${P}^{Cal}$ as follows:

\begin{equation}
\begin{aligned}
    \arg\max_{y\in K} {P}^{Cal}(y\mid x)  = \arg\max_{y\in K} {P}(x\mid y) & \\ = \arg \max_{y\in K} \{{{P}(y \mid x)}/{P(y)}\}.
\end{aligned}
\label{eq1}
\end{equation}

Since softmax cross-entropy loss indicates that ${{P}(y \mid x)}\propto e^{f_y(x)}$, then Equation~\ref{eq1} can be rewritten as:
\begin{equation}
\arg\max_{y\in K} {P}^{Cal}(y\mid x) = \arg \max_{y\in K} \{{f_y(x)}-\log \gamma _y\},
\label{eq3}
\end{equation}
where $\gamma_y$ is the estimate of the class prior ${P}(y)$.  This formulation inspires us to calibrate the logits~\footnote{Logits denotes the output of the last classification layer and the input to softmax.} before softmax cross-entropy according to the probability of occurrence of each class. In other words, we should encourage the logits of minority classes to minus a relative larger value. Inspired by Equation~\ref{eq3}, we calibrate the logits for each class before softmax cross-entropy, then the modified  cross-entropy loss can be formulated as:
\begin{equation}
\mathcal{L}_{\text {Cal}}(y; f(x))=-\log\frac{1}{\sum_{i\neq y} e^{-f_{y}(x)+f_{i}(x)+\Delta_{(y,i)}}},
\label{eq2}
\end{equation}
where $\Delta_{(y,i)}=\log(\frac{\gamma_{i}}{\gamma_y})$. For more insight into $\Delta_{(y,i)}$, it can be viewed as a pairwise label margin, which represents the desired gap between scores for $y$ and $i$. 
With this optimization objective, we aim to find the optimal pairwise label margin $\Delta_{(y,i)}$ and train the local model with our calibrated loss as usual even with label distribution skew.
\paragraph{Fine-grained Calibrated Cross-Entropy} Compared to the standard softmax cross-entropy, Equation~\ref{eq2} applies a pairwise label margin $\Delta_{(y,i)}$  to each logit.
Adjusting the value of $\Delta_{(y,i)}$ for each class is the key factor in our modified loss function. For label skewed data, motivated by the interesting idea in~\cite{cao2019learning}, we provide the following optimal pairwise label margins to minimize the test error:
 
\begin{theorem} \label{secondTheorem}
For any given input $(x,y)$, the margin of label $y$ is $d_y=f_y(x) - \max \limits_{i \neq y} f_i(x)$, which denotes the minimum distance of the data in class $y$ to the decision boundary. We show that the test error for label skewed data is bounded by $    \frac{1}{d_{y} \sqrt{n_{y}}}+\frac{1}{d_{i} \sqrt{n_{i}}}.$ 
The optimal pairwise label margin is:
\begin{equation}
  \Delta_{(y,i)}={\tau}\cdot({n_y^{-1/4}} -{n_i^{-1/4}}),
\end{equation}
 where $n_y$ and $n_i$ are the sample size of class $y$ and $i$, respectively. And $\tau$ is a hyper-parameter. 
\end{theorem}
Based on above analysis, we propose a fine-grained loss function for local training that calibrates the logits based on the enforced pairwise label margins to reduce the bias in local update:
\begin{equation}\label{calibratedBound}
\mathcal{L}_{\text {Cal}}(y; f(x))=-\log\frac{e^{f_{y}(x)-\tau\cdot n_y^{-1/4}}}{\sum_{i\neq y} e^{f_{i}(x)-\tau\cdot n_{i}^{-1/4}}}.
\end{equation}
This loss function simultaneously minimizes the classification errors and forces the learning to focus on margins of minority classes to reach the optimal results. During local training, $\mathcal{L}_{\text {Cal}}$ should give an optimal trade-off between the margins of classes.
\paragraph{Deviation Bound after Calibration} In this paragraph, we demonstrate that FedLC can mitigate the deviation of local gradient as follows:
\begin{theorem}\label{thirdTheorem}
For majority class $j \in \mathcal{J}$ and minority class $r \in \mathcal{R}$, after adding the pairwise label margin $\Delta_{(y,i)}=\tau({n_y^{-1/4}}-n_{i}^{-1/4})$ for all pairs of classes, the deviation bound becomes:
$$
D_{jr}=\sum_{\substack{y \in K, y \neq r}   }\Delta_{(r,y)}\frac{\overline{\widetilde{p}_{y}^{(r)}}||\overline{h^{(r)}}||_2^2}{\overline{\widetilde{p}^{(j)}_{r}}\overline{h^{(r)}}\cdot\overline{h^{(j)}}}$$ where $\overline{\widetilde{p}_{r}^{(j)}}=\frac{1}{n_j}\sum_{i\in O_j} \widetilde{p}_{r}(x_{i})$ is averaged over all samples in majority class $j$, $\overline{\widetilde{p}_{y}^{(r)}}=\frac{1}{n_r}\sum_{i\in O_r} \widetilde{p}_{y}(x_{i})$ is averaged over all samples in majority class $r$. And also we have $\widetilde{p}_{r}(x)=\frac{e^{f_r(x)}}{\sum_{i=1}^K e^{f_{i}(x)-\tau\cdot n_{i}^{-1/4}}}$ , $\widetilde{p}_{y}(x)=\frac{e^{f_y(x)}}{\sum_{i=1}^K e^{f_{i}(x)-\tau\cdot n_{i}^{-1/4}}}$. 
\end{theorem}

Obviously, the modified loss adds  pairwise label margin directly into the deviation bound to enlarge $D_{jr}$ for minority class $r$ and make $n_{j}/n_{r} \gg D_{jr}$ more difficult, which mitigate the deviation of $\{\Delta w_y\}_{y=1}^K$ during local training. Thus in this way, we can reduce the bias of local model updates for each client, which in turn benefits the global model.





\section{Experiments} 


\subsection{Type of Label Distribution Skew}
\label{statement} 
\begin{figure}[t] 
    \centering
    \includegraphics[width=8cm]{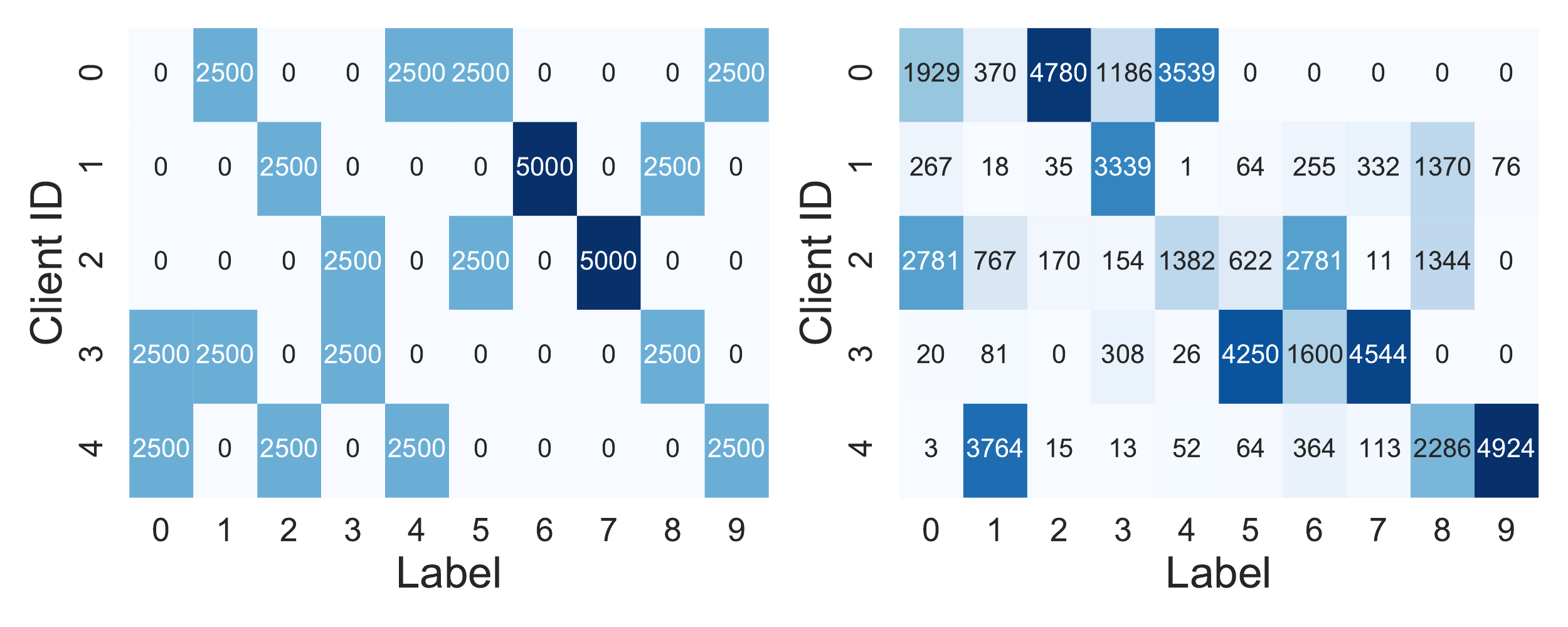}
    \caption{Visualizations of skewed CIFAR10 dataset on 5 clients. \textbf{Left:} quantity-based label skew ($\alpha=4$); \textbf{Right:} distribution-based label skew ($\beta=0.5$). The value in each rectangle is the number of data samples of a label belonging to a certain client.
    }
    \label{distribution}
\end{figure}

To simulate label distribution skew, we follow the settings in~\cite{li2021federated} and introduce two frequently used label skew settings: quantity-based label skew and distribution-based label skew.  An example of different types of label distribution skew is shown in Figure~\ref{distribution}.

\paragraph{Quantity-based Label skew} It is first introduced in FedAvg~\cite{mcmahan2017communication}, and has been frequently used in many recent studies~\cite{li2021fedrs, DBLP:conf/mlsys/LiSZSTS20,DBLP:journals/corr/abs-2006-16765,DBLP:conf/aaai/LiWH20}.  Suppose that there are $n$ training samples distributed among $m$ clients. Firstly, we sort the data by labels and divide it into $m \cdot \alpha$ sets, each set contains $\frac{n}{m\cdot \alpha}$ samples. Then we assign $\alpha$ sets to each client.  We refer to this approach as $Q(\alpha)$, where $\alpha$ controls the degree of label skew. Note that there is no overlap between the samples of different clients. Each client's training data contains only a few labels, which means there are missing classes.
\begin{table*}[t]
\centering
    \caption{Performance overview for different  degrees of distribution-based label skew. }
    \label{skew_distribution}
    \scalebox{0.67}{
    \begin{tabular}{c|cccc|cccc|cccc}
    \toprule
    Dataset & \multicolumn{4}{c|}{SVHN} & \multicolumn{4}{c|}{CIFAR10} & \multicolumn{4}{c}{CIFAR100} \\
    \midrule
    Skewness & $\beta=0.05$ & $\beta=0.1$ & $\beta=0.3$ & $\beta=0.5$ & $\beta=0.05$ & $\beta=0.1$ & $\beta=0.3$ & $\beta=0.5$ & $\beta=0.05$ & $\beta=0.1$ & $\beta=0.3$ & $\beta=0.5$ \\
    \midrule
    FedAvg & $69.51_{\pm1.45}$ & $79.86_{\pm1.46}$ & $85.14_{\pm0.83}$ & $86.02_{\pm1.15}$ & $37.63_{\pm1.36}$ & $48.07_{\pm1.38}$ & $55.95_{\pm0.83}$ & $60.18_{\pm1.78}$ & $21.37_{\pm0.87}$ & $25.06_{\pm1.04}$ & $28.44_{\pm1.51}$ & $29.29_{\pm1.32}$ \\
    FedProx & $71.42_{\pm1.24}$ & $81.39_{\pm1.35}$ & $86.30_{\pm0.95}$ & $87.53_{\pm1.56}$ & $39.03_{\pm1.27}$ & $49.57_{\pm0.90}$ & $57.88_{\pm0.93}$ & $62.13_{\pm1.17}$ & $22.92_{\pm1.71}$ & $26.44_{\pm0.86}$ & $30.16_{\pm1.18}$ & $31.20_{\pm1.23}$ \\
    Scaffold & $71.23_{\pm1.63}$ & $81.80_{\pm1.75}$ & $86.32_{\pm1.19}$ & $87.13_{\pm1.39}$ & $38.84_{\pm0.93}$ & $49.12_{\pm1.21}$ & $57.39_{\pm1.16}$ & $61.54_{\pm1.28}$ & $22.61_{\pm1.37}$ & $26.30_{\pm1.32}$ & $29.96_{\pm1.17}$ & $31.26_{\pm1.75}$ \\
    FedNova & $72.50_{\pm1.21}$ & $82.41_{\pm1.40}$ & $87.11_{\pm1.38}$ & $86.65_{\pm1.25}$ & $39.81_{\pm1.18}$ & $50.56_{\pm1.42}$ & $58.85_{\pm0.93}$ & $62.77_{\pm0.86}$ & $24.03_{\pm0.91}$ & $27.65_{\pm0.99}$ & $30.76_{\pm0.95}$ & $31.93_{\pm0.98}$ \\
    FedOpt & $73.46_{\pm1.07}$ & $82.71_{\pm1.13}$ & $86.85_{\pm0.85}$ & $87.41_{\pm1.72}$ & $41.08_{\pm1.01}$ & $51.89_{\pm0.86}$ & $59.39_{\pm1.68}$ & $63.38_{\pm1.62}$ & $24.51_{\pm1.71}$ & $28.98_{\pm1.08}$ & $32.42_{\pm1.66}$ & $32.94_{\pm1.28}$ \\
    FedRS & $75.97_{\pm1.15}$ & $83.27_{\pm1.54}$ & $87.01_{\pm0.98}$ & $87.40_{\pm1.67}$ & $44.39_{\pm1.63}$ & $54.04_{\pm1.59}$ & $62.40_{\pm1.38}$ & $66.39_{\pm1.28}$ & $27.93_{\pm1.18}$ & $32.89_{\pm1.50}$ & $36.58_{\pm0.94}$ & $38.98_{\pm1.35}$ \\
    \midrule
    Ours & $\textbf{82.36}_{\pm0.67}$ & $\textbf{84.41}_{\pm0.87}$ & $\textbf{88.02}_{\pm1.19}$ & $ \textbf{88.48}_{\pm1.29} $ &  $\textbf{54.55}_{\pm1.70} $ & $ \textbf{65.91}_{\pm1.68} $ & $ \textbf{72.18}_{\pm0.86} $ & $ \textbf{72.99}_{\pm1.12} $ &  $\textbf{38.08}_{\pm0.84} $ & $ \textbf{41.01}_{\pm1.08} $ & $ \textbf{44.23}_{\pm1.70} $ & $ \textbf{44.96}_{\pm1.71}$ \\
    \bottomrule
    \end{tabular}}
\end{table*}

\begin{table*}[t]
\centering
\caption{Performance overview for different  non-IID settings on CIFAR10 and CIFAR100 (quantity-based label skew). }
\label{quantity_skew}
\scalebox{0.9}{
    \begin{tabular}{c|cccc|cccc}
        \toprule
    Dataset & \multicolumn{4}{c|}{CIFAR10} & \multicolumn{4}{c}{CIFAR100} \\
    \midrule
    Degree of skewness & $\alpha=2$ & $\alpha=4$ & $\alpha=6$ & $\alpha=8$ & $\alpha=20$ & $\alpha=40$ & $\alpha=60$ & $\alpha=80$ \\
    \midrule
    FedAvg & $52.23_{\pm1.01}$ & $71.29_{\pm1.49}$ & $77.96_{\pm0.92}$ & $78.49_{\pm1.30}$ & $42.21_{\pm1.38}$ & $47.61_{\pm1.20}$ & $49.54_{\pm1.60}$ & $49.81_{\pm1.84}$ \\
    FedProx & $52.97_{\pm1.12}$ & $72.51_{\pm1.17}$ & $78.74_{\pm1.55}$ & $80.16_{\pm0.92}$ & $43.33_{\pm1.40}$ & $48.53_{\pm1.64}$ & $51.16_{\pm1.72}$ & $51.25_{\pm1.28}$ \\
    Scaffold & $53.85_{\pm1.75}$ & $72.33_{\pm1.04}$ & $78.89_{\pm1.38}$ & $79.88_{\pm1.65}$ & $43.69_{\pm1.07}$ & $49.26_{\pm1.52}$ & $50.95_{\pm1.24}$ & $51.42_{\pm0.85}$ \\
    FedNova & $53.27_{\pm1.40}$ & $73.87_{\pm1.44}$ & $80.18_{\pm1.03}$ & $79.70_{\pm1.59}$ & $43.14_{\pm1.35}$ & $49.39_{\pm1.69}$ & $51.02_{\pm1.02}$ & $51.51_{\pm0.92}$ \\
    FedOpt & $54.35_{\pm1.29}$ & $72.51_{\pm1.52}$ & $79.47_{\pm1.46}$ & $80.05_{\pm1.28}$ & $43.97_{\pm1.39}$ & $49.07_{\pm1.13}$ & $50.98_{\pm0.96}$ & $51.27_{\pm1.95}$ \\
    FedRS & $55.91_{\pm1.44}$ & $74.44_{\pm1.15}$ & $80.62_{\pm1.13}$ & $80.29_{\pm1.02}$ & $45.58_{\pm1.90}$ & $49.95_{\pm1.76}$ & $50.83_{\pm1.16}$ & $51.99_{\pm1.51}$ \\
    \midrule
    Ours & $\textbf{61.87}_{\pm0.94} $ & $ \textbf{77.22}_{\pm1.22} $ & $ \textbf{81.41}_{\pm1.18} $ & $ \textbf{82.05}_{\pm0.89} $ & $ \textbf{47.77}_{\pm0.84} $ & $ \textbf{52.26}_{\pm1.43} $ & $ \textbf{52.89}_{\pm1.34} $ & $ \textbf{53.42}_{\pm0.85}$ \\
    \bottomrule
    \end{tabular}}
\end{table*}

\paragraph{Distribution-based Label skew} This
partitioning strategy was first introduced in~\cite{yurochkin2019bayesian}, such a setting is also used in many other studies~\cite{DBLP:conf/ijcai/LiHS21,DBLP:conf/nips/LinKSJ20,DBLP:conf/iclr/WangYSPK20,zhang2021practical}. Each client is allocated a proportion of the samples of each label according to Dirichlet distribution.  In detail, we sample the data by simulating $\mathbf{p}_{k} \sim \operatorname{Dir}(\beta)$ and allocate a portion of $\mathbf{p}_{k,j}$ of the samples in class $k$ to client $j$.
For ease of presentation, we use $D(\beta)$ to denote such a partitioning strategy. Here $\beta$ controls the degree of skewness.  Note that when using this partitioning strategy, the training data of each client may have majority classes, minority classes, or even some missing classes, which is more practical in real-world applications. 


\subsection{Experimental Setups}

\begin{table}[t]
\centering
\caption{Performance overview given different skewed ImageNet-subset datasets.}
\label{imagenet_sub}
\scalebox{0.9}{
\begin{tabular}{c|cc|cc}
\toprule
Degree of skewness & $\alpha=2$ & $\alpha=4$ & $\beta=0.1$ & $\beta=0.3$ \\
\midrule
FedAvg & 53.87 & 63.12 & 43.91 & 56.21 \\
FedProx & 55.87 & 65.19 & 45.75 & 58.13 \\
Scaffold & 56.27 & 64.64 & 46.34 & 57.92 \\
FedNova & 56.02 & 65.29 & 46.05 & 58.45 \\
FedOpt & 55.91 & 65.17 & 45.89 & 58.08 \\
FedRS & 57.23 & 67.78 & 47.23 & 59.83 \\
\midrule
Ours & \textbf{62.43} & \textbf{72.33} & \textbf{53.24} & \textbf{63.81} \\
\bottomrule
\end{tabular}}
\end{table}

\paragraph{Datasets} 
In this study, we conduct a number of experiments on popular image classification benchmark datasets: SVHN, CIFAR10, CIFAR100 and ImageNet~\cite{DBLP:conf/cvpr/DengDSLL009}, as well as federated datasets (Synthetic dataset and FEMNIST) proposed in LEAF~\cite{caldas2019leaf}. According to~\cite{li2021anti}, we generate the Imagenet-subset with size 64*64*3 , which consists of 12 labels for fast training. 
To better simulate label distribution skew in Synthetic, a mixed manner consists of quantity-based label skew and distribution-based label skew are used in our experiments.  We denote it as Synthetic$(\lambda,\mu)$, where local data size follows a power law. Note that $\lambda$ specifies how much local models differ from one another, and $\mu$ indicates how skewed the local data is at each 
client. We use a simple logistic model ($y=\operatorname{\arg\max}(\operatorname{softmax}(W x+b))$) to generate data samples. For FEMNIST, we use the default setting in LEAF~\cite{caldas2019leaf}. 

\paragraph{Baselines and Implementation} Our method aims at improving the performance of federated learning with label distribution skew. As a result, we choose typical approaches to non-IID issues as our baselines, such as FedProx~\cite{li2018federated},Scaffold~\cite{karimireddy2020scaffold}, FedNova~\cite{DBLP:conf/nips/WangLLJP20} and FedOpt~\cite{DBLP:conf/iclr/ReddiCZGRKKM21} as our baselines. For fair comparison, we also compare our method with FedRS~\cite{li2021fedrs}, which focuses on the issue of label skew in federated learning.
We implement the typical federated setting~\cite{mcmahan2017communication} in Pytorch, and all experiments are conducted with 8 Tesla V100 GPUs. At default, there are 20 clients totally.  The size of local mini-batch is 128. For local training, each client updates the weights via SGD optimizer with learning rate $\eta=0.01$ without weight decay. We run each experiment with 5 random seeds and report the average and standard deviation.

   

\subsection{Experiments on Federated Datasets}
\begin{table}[t]
    \centering
    \caption{Test accuracy on various federated datasets. }
    \label{tb_syn}
    \scalebox{0.72}{
    \begin{tabular}{c|ccc|c}
    \toprule
    Dataset & Synthetic(0,0)      & Synthetic(0.5,0.5)  & Synthetic(1,1)      & FEMNIST             \\
    \midrule
    FedAvg  & $72.09_{\pm1.19}$ & $67.55_{\pm1.16}$ & $63.11_{\pm0.66}$ & $84.14_{\pm0.73}$ \\
    FedProx & $72.21_{\pm0.47} $ & $ 67.85_{\pm0.82} $ & $ 63.09_{\pm0.79} $ & $ 85.41_{\pm1.28}$ \\
    Scaffold & $73.51_{\pm0.67} $ & $ 69.45_{\pm0.82} $ & $ 64.95_{\pm0.83} $ & $ 86.24_{\pm1.29}$ \\
    FedNova & $73.22_{\pm0.94} $ & $ 69.01_{\pm0.87} $ & $ 64.67_{\pm0.46} $ & $ 85.32_{\pm1.62}$ \\
    FedOPT  & $73.42_{\pm0.76} $ & $ 68.64_{\pm0.67} $ & $ 63.69_{\pm0.89} $ & $ 86.56_{\pm0.48}$ \\
    FedRS   & $75.79_{\pm1.25} $ & $ 71.35_{\pm0.81} $ & $ 66.39_{\pm1.41} $ & $ 87.95_{\pm0.92}$ \\
    \midrule
    Ours    & $\textbf{80.92}_{\pm0.31} $ & $ \textbf{78.47}_{\pm0.34} $ & $ \textbf{75.46}_{\pm0.52} $ & $ \textbf{92.78}_{\pm0.58}$ \\
    \bottomrule
    \end{tabular}}
\end{table}
In this section, we evaluate these algorithms on Synthetic and FEMNIST dataset. We divide the training data into 100 clients over 300 communication rounds. To manipulate heterogeneity more precisely, we synthesize unbalanced datasets with 3 different settings. Specifically, we follow the settings in~\cite{li2018federated} and generate Synthetic(0,0), Synthetic(0.5,0.5), and Synthetic(1,1). 

Observed from Table~\ref{tb_syn}, we find that a large $\lambda$ and $\mu$ on the Synthetic datasets can lead to poor performance on test accuracy. Especially, for Synthetic(1,1), our method achieves a prediction
accuracy of 75.45\%, which is higher than that of the best baseline FedRS by 9.07\%. 
 Obviously, we can find that on all datasets, our method consistently outperforms much better than any other baseline. As a point of interest, more similarities in performance can be observed between these baselines, which indicates that these methods are not appropriate for highly skewed data distribution.
\subsection{Experiments on Real-World Datasets}
\label{sec_exp}
\paragraph{Results on SVHN, CIFAR10 and CIFAR100} All results are tested after 400 communication rounds. We mainly report the evaluations of these algorithms with different degrees of label skew. Table~\ref{skew_distribution} summarizes the results for different types of distribution-based label skew. Evidently, in all scenarios, our method significantly achieves a higher accuracy than other SOTA methods. As data heterogeneity increases (i.e.\ ,  smaller $\beta$), all competing methods struggle, whereas our method displays markedly improved 
accuracy on highly skewed data. For example, for
CIFAR-10 dataset with $\beta=0.05$, our method gets a test accuracy of 54.55\%,
which is much higher than that of FedRS by 10.16\%. In addition, we also report the performance of these methods for different types of quantity-based label skew in Table~\ref{quantity_skew}, which can further demonstrate the superiority of our method.

\begin{figure}[t] 
    \centering
    \includegraphics[width=7.5cm]{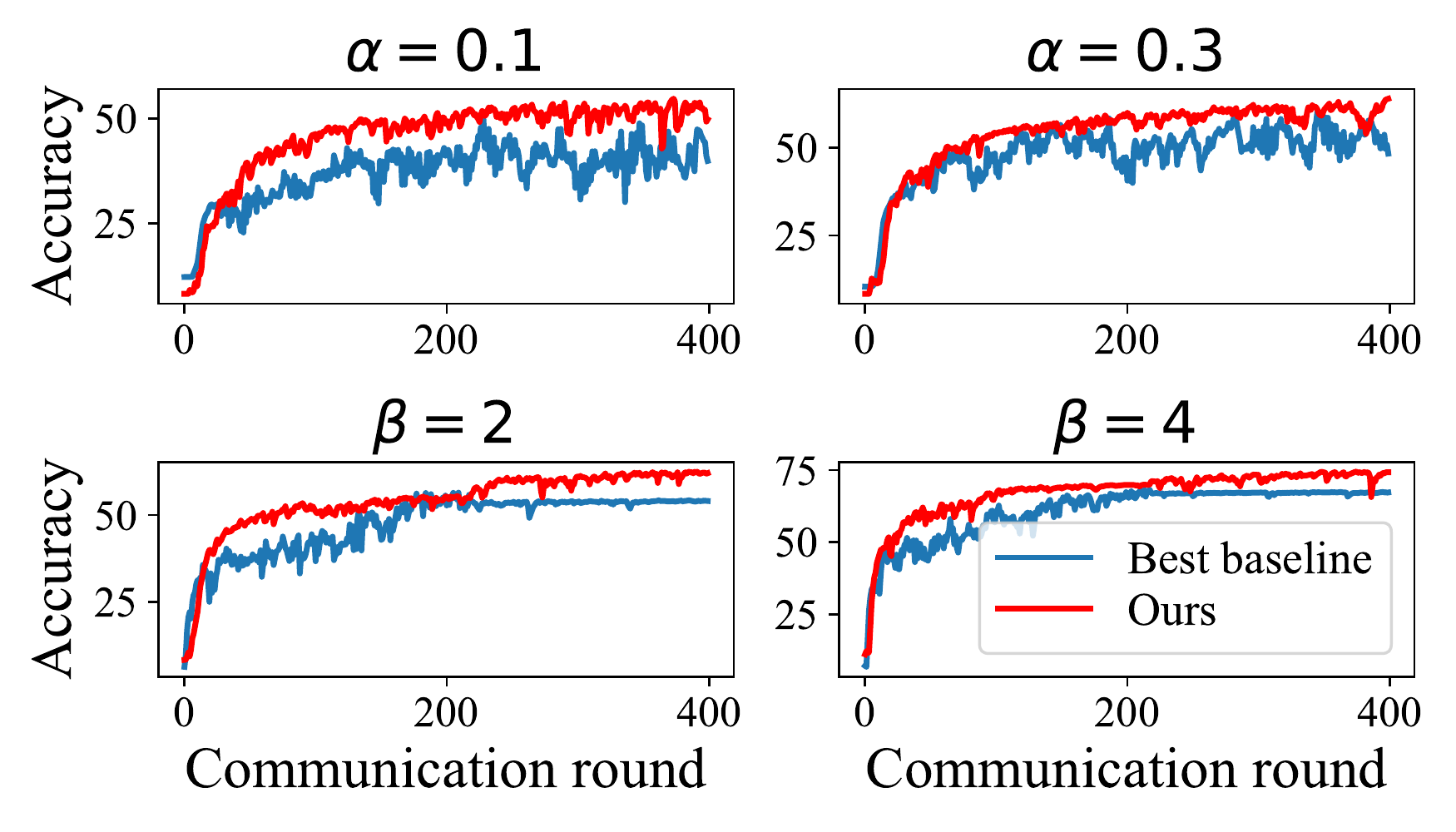}
    \caption{Test accuracy for different skewed ImageNet-subset datasets. We compare our method with the best performed baseline FedRS on 40 clients.
    }
    \label{imgnet_plot}
\end{figure}
\begin{figure}[t]  
    \centering 
    \includegraphics[width=8cm]{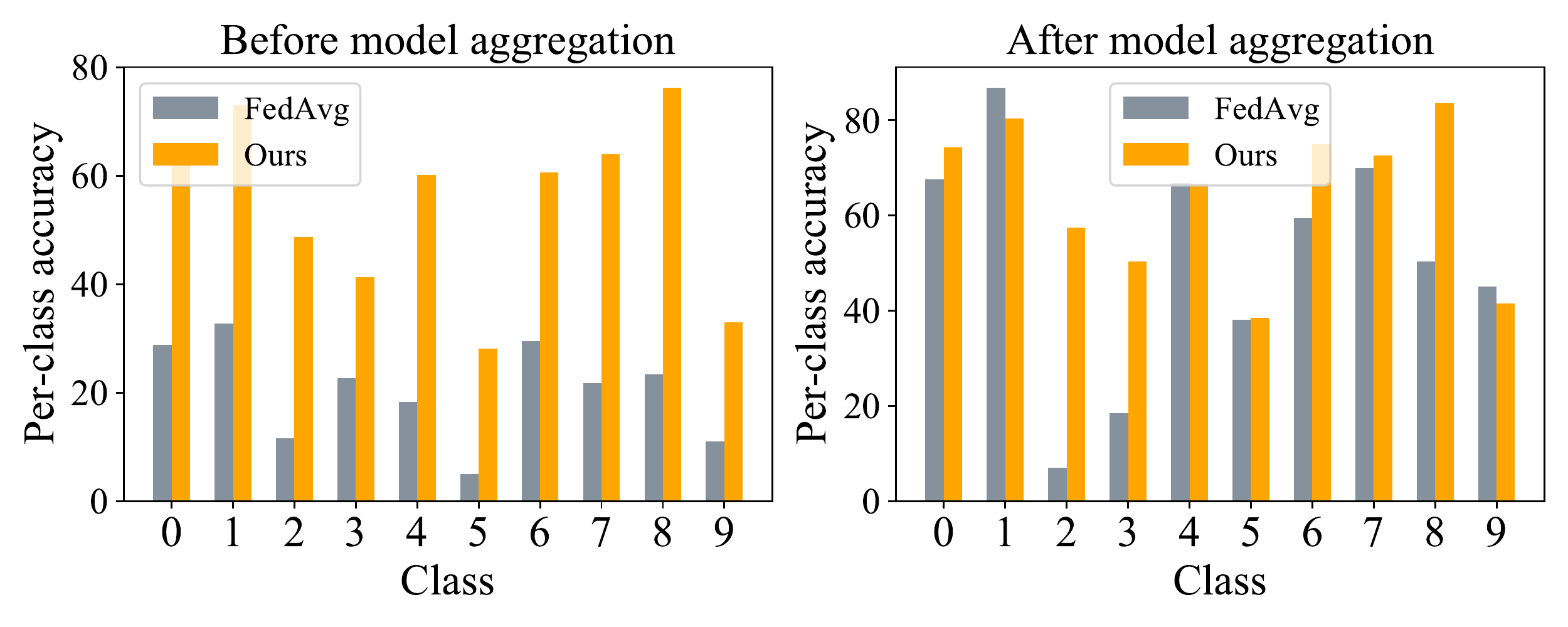}
    \caption{Average per-class accuracy before and after model aggregation. For fair comparisons, we use the same well-trained model for initialization and the same data partition on each client.}
    \label{avg_acc}
    \vspace{-0.4cm}
\end{figure}
\begin{figure}[t] 
    \centering
    \includegraphics[width=8cm]{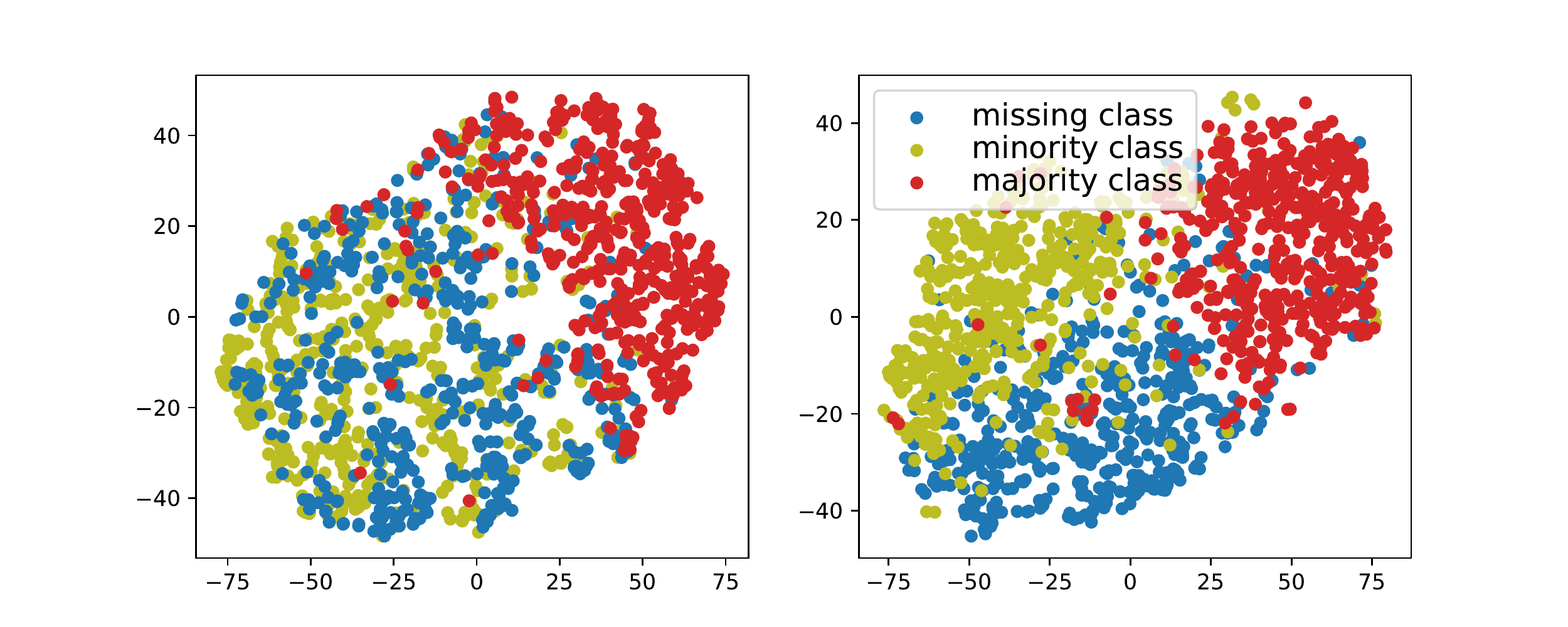}
    \caption{TSNE visualizations on majority, minority and missing classes. \textbf{Left: }For FedAvg, the samples from the minority class and missing class are mixed together and indistinguishable. \textbf{Right: } For our method, the data from minority class and missing class can be distinguished well, which indicates our method can learn more discriminative features.
}
    \label{tsne}
    \vspace{-0.5cm}
\end{figure}

\paragraph{Results on ImageNet-subset} In addition, we compare the prediction performance of these methods on ImageNet-subset dataset. We choose ResNet-18 as the default network. We compare these FL algorithms in terms of both quantity-based label skew and distribution-based label skew with $\alpha=\{2,4\}$ and $\beta=\{0.1,0.3\}$. As shown in Table~\ref{imagenet_sub}, we have to emphasize here that our method still performs better even on such a complex dataset. In the case of highly skewed data with $\alpha=2$ and $\beta=0.1$, our method can achieve a mean accuracy of 62.43\% and 54.43\%,  which is better than the best baseline FedRS by 7.2\% and 6.01\%, respectively. Besides, we plot the 
test accuracy curves in Figure~\ref{imgnet_plot}, which means that our method is relatively stable on highly skewed data (e.g.\ , $\alpha=0.1$). 
\begin{table}[t]
\centering
\caption{Performance overview on CIFAR10 and SVHN for different local epochs $E$. All experiments are conducted on 10 clients.}
\label{local_e}
\scalebox{0.65}{
\begin{tabular}{c|c|cc|cc}
\toprule
 & Dataset & \multicolumn{2}{c}{CIFAR10} & \multicolumn{2}{c}{SVHN} \\
 \midrule
\# Local Epoch & Skewness & $\beta=0.5$ & $\alpha=5$ & $\beta=0.5$ & $\alpha=5$ \\
 \midrule
\multirow{4}{*}{E=1} & FedAvg & 54.25 & 46.12 & 82.07 & 84.91 \\
 & FedProx & 54.68 & 46.84 & 83.18 & 85.22 \\
 & FedNova & 56.14 & 47.58 & 83.24 & 85.14 \\
 & Ours & \textbf{70.41} & \textbf{58.43} & \textbf{85.89} & \textbf{88.34} \\
  \midrule
\multirow{4}{*}{E=5} & FedAvg & 77.02 & 75.81 & 89.01 & 88.59 \\
 & FedProx & 77.48 & 75.14 & 90.01 & 89.12 \\
 & FedNova & 78.64 & 75.88 & 90.17 & 88.79 \\
 & Ours & \textbf{80.25} & \textbf{77.76} & \textbf{91.83} & \textbf{91.25} \\
 \midrule
\multirow{4}{*}{E=10} & FedAvg & 80.07 & 78.81 & 87.82 & 89.15 \\
 & FedProx & 80.25 & 78.72 & 88.18 & 89.45 \\
 & FedNova & 80.56 & 78.97 & 88.16 & 89.53 \\
 & Ours & \textbf{81.67} & \textbf{79.61} & \textbf{89.54} & \textbf{90.56} \\
  \midrule
\multirow{4}{*}{E=20} & FedAvg & 80.61 & 79.21 & 87.39 & 89.42 \\
 & FedProx & 80.17 & 79.13 & 88.01 & 89.14 \\
 & FedNova & 80.45 & 79.32 & 87.52 & 89.24 \\
 & Ours & \textbf{81.99} & \textbf{80.19} & \textbf{88.75} & \textbf{89.46} \\
 \bottomrule
\end{tabular}}
\end{table}
\begin{table}[t]
\centering
\caption{Test accuracy with different number of clients. }
\label{clients}
\scalebox{0.8}{
\begin{tabular}{c|cccc}
\toprule
\# Clients & $m=10$ & $m=30$ & $m=50$ & $m=100$ \\
\midrule
FedAvg & 57.19 & 45.09 & 36.70 & 29.30 \\
FedProx & 59.68 & 47.11 & 39.07 & 31.74 \\
Scaffold & 59.59 & 47.14 & 38.20 & 31.09 \\
FedNova & 58.72 & 47.49 & 38.74 & 33.30 \\
FedOpt & 61.25 & 48.19 & 40.98 & 33.40 \\
FedRS & 62.62 & 50.05 & 41.98 & 35.95 \\
\midrule
Ours & \textbf{70.13} & \textbf{58.79} & \textbf{52.08} & \textbf{43.81} \\
\bottomrule
\end{tabular}}
\vspace{-0.4cm}
\end{table}

\subsection{Analyse of our method} 

\paragraph{FedMC can mitigate over-fitting} To verify the effectiveness of our method, we compare the average per-class accuracy of our method with FedAvg before and after model aggregation. For fair comparison, we use a same well-trained federated model as the global model. Then the central server  distributes the model parameter to all clients. Next, we use FedAvg and our method to train the local model with the same local data for only 1 epoch. At last, we report the average per-class accuracy of all clients. Note that there is only one difference in the whole process - the local update.
According to the left sub-figure of Figure~\ref{avg_acc},  our method's average per-class accuracy is much higher than that of FedAvg. As we mentioned before(see Figure~\
 \ref{overfit_fig}), for FedAvg, due to the highly skewed label distribution, each client has very low accuracy on minority classes and missing classes. Therefore, after the local update, each client's model is biased, resulting in a lower average accuracy for each class.  By contrast, with our method, each client has higher performance in all classes. The results show that the our new cross-entropy loss can indeed improve the performance on minority classes and missing classes, and further improve the overall performance of the global model. As shown in the right sub-figure of Figure~\ref{avg_acc}, our method can also perform better than FedAvg after model aggregation, which means our method alleviates the over-fitting in local update.

\paragraph{T-SNE visualization on majority, minority and missing classes} Additionally, we also show a t-SNE~\cite{van2008visualizing} visualization in Figure~\ref{tsne} after local update. The number of training data is $\{0,448,2608\}$ for missing, minority and  majority class. As illustrated in this figure, the test data from different classes are hard to be separated by FedAvg, only the features of the majority class are obvious, minority class and missing class are mixed together and difficult to distinguish.  Actually, our method can learn more discriminative features, which indicates that our method certainly yields better performance. 

\subsection{Ablation Study}
\paragraph{Performance for different local epochs $E$} 
In this section, we add more computation per client on each round by increasing the local epochs $E$. We conduct many experiments on CIFAR10 and SVHN dataset. There are 20 clients totally. We compare these FL algorithms in terms of both quantity-based label skew and distribution-based label skew with $\alpha=5$ and $\beta=0.5$. As shown in Table~\ref{local_e}, in cases where $E$=1, our method is superior to other baselines by a large margin. The model is not well trained when the number of local updates is too small. Our method aims to reduce such bias in local updates. For a large value of $E$, while the performance of these baselines is relatively similar, our approach still consistently outperforms other methods. 


\paragraph{Performance for different number  of  clients $m$ } To show the effectiveness  of FedLC, we train these methods with different numbers of clients $m$.
Table~\ref{clients} reports all the results across $m=\{10,30, 50, 100\}$.
As expected, our proposed FedLC achieves the best performance across all settings, which further validates that FedLC can be applied in most practical settings. 
As $m$ increases, the performance of all methods decreases. We conjecture that the reason is that 
more clients in FL make the model harder to converge. However, our method can still achieve 43.81\% accuracy when there are 100 clients.

\paragraph{Combination with other techniques}
As we focus on addressing the intra-client skew, we believe combining effective methods that target inter-client objective inconsistency with our approach can further improve the performance of the server model.
In this section, we integrate additional regularization terms used in FedNova and FedProx into our method. As illustrated in Table~\ref{final}, the combination of Scaffold and FedProx with our method leads to a better performance.
We would like to argue that this is reasonable.  Regularizing the inconsistent objectives can prevent local models from updating towards their local minima. Based on our analysis, we believe that incorporating approaches focused on intra-client label skew and inter-client objective inconsistency will result in more efficient and effective FL algorithms, which we leave as future work.  



\begin{table}[t]
    \centering
    \caption{Performance overview on the combination of Scaffold and FedProx with our method.}
    \scalebox{0.8}{
    \begin{tabular}{c|cc|cc}
        \toprule
    dataset & \multicolumn{2}{c|}{CIFAR10} & \multicolumn{2}{c}{SVHN} \\
    \midrule
    Degree of skewness & $\alpha=5$ & $\beta=0.3$ & $\alpha=5$ & $\beta=0.3$ \\
    \midrule
    FedAvg & 61.54 & 60.18 & 84.95 & 84.85 \\
    \midrule
    Ours & 69.89 & 70.38 & 88.19 & 88.31 \\
    \midrule
    Ours+Scaffold & 70.45 & 71.47 & 89.34 & 89.45 \\
    \midrule
    Ours+FedProx & 71.39 & 71.98 & 90.32 & 89.49 \\
    \bottomrule
    \end{tabular}}
    \label{final}
    \vspace{-0.5cm}
\end{table}

\section{Conclusion}
In this work, we propose a fine-grained calibrated loss to improve the performance of the global model with label distribution skew. The comprehensive experiments 
demonstrate the proposed method can effectively reduce the bias in local update. We hope that our study can inspire other researchers to explore more connections between  intra-client label skew and inter-client objective inconsistency. 

\section{Acknowledgment}
This work was supported by the National Key Research and Development Project of China (2021ZD0110400 \& No. 2018AAA0101900), National Natural Science Foundation of China (U19B2042), The University Synergy Innovation Program of Anhui Province (GXXT-2021-004), Zhejiang Lab (2021KE0AC02), Academy Of Social Governance Zhejiang University, Fundamental Research Funds for the Central Universities (226-2022-00064), Artificial Intelligence Research Foundation of Baidu Inc., Program of ZJU and Tongdun Joint Research Lab.

{\small
\bibliographystyle{ieee_fullname}
\bibliography{egbib}

\begin{thebibliography}{10}\itemsep=-1pt

\bibitem{acar2021federated}
Durmus Alp~Emre Acar, Yue Zhao, Ramon Matas, Matthew Mattina, Paul Whatmough,
  and Venkatesh Saligrama.
\newblock Federated learning based on dynamic regularization.
\newblock In {\em International Conference on Learning Representations}, 2021.

\bibitem{alshedivat2021federated}
Maruan Al-Shedivat, Jennifer Gillenwater, Eric Xing, and Afshin Rostamizadeh.
\newblock Federated learning via posterior averaging: A new perspective and
  practical algorithms, 2021.

\bibitem{caldas2019leaf}
Sebastian Caldas, Sai Meher~Karthik Duddu, Peter Wu, Tian Li, Jakub Konečný,
  H.~Brendan McMahan, Virginia Smith, and Ameet Talwalkar.
\newblock Leaf: A benchmark for federated settings, 2019.

\bibitem{cao2019learning}
Kaidi Cao, Colin Wei, Adrien Gaidon, Nikos Arechiga, and Tengyu Ma.
\newblock Learning imbalanced datasets with label-distribution-aware margin
  loss.
\newblock {\em arXiv preprint arXiv:1906.07413}, 2019.

\bibitem{DBLP:conf/aistats/CharlesK21}
Zachary Charles and Jakub Kone{\v{c}}n{\'y}.
\newblock Convergence and accuracy trade-offs in federated learning and
  meta-learning.
\newblock In Arindam Banerjee and Kenji Fukumizu, editors, {\em The 24th
  International Conference on Artificial Intelligence and Statistics, {AISTATS}
  2021, April 13-15, 2021, Virtual Event}, volume 130 of {\em Proceedings of
  Machine Learning Research}, pages 2575--2583. {PMLR}, 2021.

\bibitem{chawla2002smote}
Nitesh~V Chawla, Kevin~W Bowyer, Lawrence~O Hall, and W~Philip Kegelmeyer.
\newblock Smote: synthetic minority over-sampling technique.
\newblock {\em Journal of artificial intelligence research}, 16:321--357, 2002.

\bibitem{Chen_2022_CVPR}
Zhaoyu Chen, Bo Li, Jianghe Xu, Shuang Wu, Shouhong Ding, and Wenqiang Zhang.
\newblock Towards practical certifiable patch defense with vision transformer.
\newblock In {\em Proceedings of the IEEE/CVF Conference on Computer Vision and
  Pattern Recognition (CVPR)}, pages 15148--15158, June 2022.

\bibitem{cui2019class}
Yin Cui, Menglin Jia, Tsung-Yi Lin, Yang Song, and Serge Belongie.
\newblock Class-balanced loss based on effective number of samples.
\newblock In {\em Proceedings of the IEEE/CVF Conference on Computer Vision and
  Pattern Recognition}, pages 9268--9277, 2019.

\bibitem{DBLP:conf/cvpr/DengDSLL009}
Jia Deng, Wei Dong, Richard Socher, Li{-}Jia Li, Kai Li, and Li Fei{-}Fei.
\newblock Imagenet: {A} large-scale hierarchical image database.
\newblock In {\em 2009 {IEEE} Computer Society Conference on Computer Vision
  and Pattern Recognition {(CVPR} 2009), 20-25 June 2009, Miami, Florida,
  {USA}}, pages 248--255. {IEEE} Computer Society, 2009.

\bibitem{he2009learning}
Haibo He and Edwardo~A Garcia.
\newblock Learning from imbalanced data.
\newblock {\em IEEE Transactions on knowledge and data engineering},
  21(9):1263--1284, 2009.

\bibitem{DBLP:conf/mmm/HuSLYL17}
Jiagao Hu, Zhengxing Sun, Bo Li, Kewei Yang, and Dongyang Li.
\newblock Online user modeling for interactive streaming image classification.
\newblock In {\em MultiMedia Modeling - 23rd International Conference, {MMM}
  2017, Reykjavik, Iceland, January 4-6, 2017, Proceedings, Part {II}}, volume
  10133 of {\em Lecture Notes in Computer Science}, pages 293--305. Springer,
  2017.

\bibitem{jamal2020rethinking}
Muhammad~Abdullah Jamal, Matthew Brown, Ming-Hsuan Yang, Liqiang Wang, and
  Boqing Gong.
\newblock Rethinking class-balanced methods for long-tailed visual recognition
  from a domain adaptation perspective.
\newblock In {\em Proceedings of the IEEE/CVF Conference on Computer Vision and
  Pattern Recognition}, pages 7610--7619, 2020.

\bibitem{kairouz2021advances}
Peter Kairouz and H.~Brendan McMahan.
\newblock Advances and open problems in federated learning, 2021.

\bibitem{kang2019decoupling}
Bingyi Kang, Saining Xie, Marcus Rohrbach, Zhicheng Yan, Albert Gordo, Jiashi
  Feng, and Yannis Kalantidis.
\newblock Decoupling representation and classifier for long-tailed recognition.
\newblock {\em arXiv preprint arXiv:1910.09217}, 2019.

\bibitem{karimireddy2020scaffold}
Sai~Praneeth Karimireddy, Satyen Kale, Mehryar Mohri, Sashank Reddi, Sebastian
  Stich, and Ananda~Theertha Suresh.
\newblock Scaffold: Stochastic controlled averaging for federated learning.
\newblock In {\em International Conference on Machine Learning}, pages
  5132--5143. PMLR, 2020.

\bibitem{kim2020adjusting}
Byungju Kim and Junmo Kim.
\newblock Adjusting decision boundary for class imbalanced learning.
\newblock {\em IEEE Access}, 8:81674--81685, 2020.

\bibitem{DBLP:conf/aaai/LiSG19}
Bo Li, Zhengxing Sun, and Yuqi Guo.
\newblock Supervae: Superpixelwise variational autoencoder for salient object
  detection.
\newblock In {\em The Thirty-Third {AAAI} Conference on Artificial
  Intelligence, {AAAI} 2019, The Thirty-First Innovative Applications of
  Artificial Intelligence Conference, {IAAI} 2019, The Ninth {AAAI} Symposium
  on Educational Advances in Artificial Intelligence, {EAAI} 2019, Honolulu,
  Hawaii, USA, January 27 - February 1, 2019}, pages 8569--8576. {AAAI} Press,
  2019.

\bibitem{DBLP:conf/iccv/LiSLWH19}
Bo Li, Zhengxing Sun, Qian Li, Yunjie Wu, and Anqi Hu.
\newblock Group-wise deep object co-segmentation with co-attention recurrent
  neural network.
\newblock In {\em 2019 {IEEE/CVF} International Conference on Computer Vision,
  {ICCV} 2019, Seoul, Korea (South), October 27 - November 2, 2019}, pages
  8518--8527. {IEEE}, 2019.

\bibitem{DBLP:conf/icassp/LiSTH19}
Bo Li, Zhengxing Sun, Lv Tang, and Anqi Hu.
\newblock Two-b-real net: Two-branch network for real-time salient object
  detection.
\newblock In {\em {IEEE} International Conference on Acoustics, Speech and
  Signal Processing, {ICASSP} 2019, Brighton, United Kingdom, May 12-17, 2019},
  pages 1662--1666. {IEEE}, 2019.

\bibitem{DBLP:conf/ijcai/0061STSS19}
Bo Li, Zhengxing Sun, Lv Tang, Yunhan Sun, and Jinlong Shi.
\newblock Detecting robust co-saliency with recurrent co-attention neural
  network.
\newblock In Sarit Kraus, editor, {\em Proceedings of the Twenty-Eighth
  International Joint Conference on Artificial Intelligence, {IJCAI} 2019,
  Macao, China, August 10-16, 2019}, pages 818--825. ijcai.org, 2019.

\bibitem{DBLP:conf/mm/LiSWL19}
Bo Li, Zhengxing Sun, Quan Wang, and Qian Li.
\newblock Co-saliency detection based on hierarchical consistency.
\newblock In Laurent Amsaleg, Benoit Huet, Martha~A. Larson, Guillaume Gravier,
  Hayley Hung, Chong{-}Wah Ngo, and Wei~Tsang Ooi, editors, {\em Proceedings of
  the 27th {ACM} International Conference on Multimedia, {MM} 2019, Nice,
  France, October 21-25, 2019}, pages 1392--1400. {ACM}, 2019.

\bibitem{DBLP:conf/mm/LiXWDLH21}
Bo Li, Jianghe Xu, Shuang Wu, Shouhong Ding, Jilin Li, and Feiyue Huang.
\newblock Detecting adversarial patch attacks through global-local consistency.
\newblock In Dawn Song, Dacheng Tao, Alan~L. Yuille, Anima Anandkumar, Aishan
  Liu, Xinyun Chen, Yingwei Li, Chaowei Xiao, Xun Yang, and Xianglong Liu,
  editors, {\em {ADVM} '21: Proceedings of the 1st International Workshop on
  Adversarial Learning for Multimedia, Virtual Event, China, 20 October 2021},
  pages 35--41. {ACM}, 2021.

\bibitem{li2021federated}
Qinbin Li, Yiqun Diao, Quan Chen, and Bingsheng He.
\newblock Federated learning on non-iid data silos: An experimental study.
\newblock {\em arXiv preprint arXiv:2102.02079}, 2021.

\bibitem{DBLP:conf/ijcai/LiHS21}
Qinbin Li, Bingsheng He, and Dawn Song.
\newblock Practical one-shot federated learning for cross-silo setting.
\newblock In Zhi{-}Hua Zhou, editor, {\em Proceedings of the Thirtieth
  International Joint Conference on Artificial Intelligence, {IJCAI} 2021,
  Virtual Event / Montreal, Canada, 19-27 August 2021}, pages 1484--1490.
  ijcai.org, 2021.

\bibitem{DBLP:conf/aaai/LiWH20}
Qinbin Li, Zeyi Wen, and Bingsheng He.
\newblock Practical federated gradient boosting decision trees.
\newblock In {\em The Thirty-Fourth {AAAI} Conference on Artificial
  Intelligence, {AAAI} 2020, The Thirty-Second Innovative Applications of
  Artificial Intelligence Conference, {IAAI} 2020, The Tenth {AAAI} Symposium
  on Educational Advances in Artificial Intelligence, {EAAI} 2020, New York,
  NY, USA, February 7-12, 2020}, pages 4642--4649. {AAAI} Press, 2020.

\bibitem{li2018federated}
Tian Li, Anit~Kumar Sahu, Manzil Zaheer, Maziar Sanjabi, Ameet Talwalkar, and
  Virginia Smith.
\newblock Federated optimization in heterogeneous networks.
\newblock {\em arXiv preprint arXiv:1812.06127}, 2018.

\bibitem{li2020feddane}
Tian Li, Anit~Kumar Sahu, Manzil Zaheer, Maziar Sanjabi, Ameet Talwalkar, and
  Virginia Smith.
\newblock Feddane: A federated newton-type method, 2020.

\bibitem{DBLP:conf/mlsys/LiSZSTS20}
Tian Li, Anit~Kumar Sahu, Manzil Zaheer, Maziar Sanjabi, Ameet Talwalkar, and
  Virginia Smith.
\newblock Federated optimization in heterogeneous networks.
\newblock In Inderjit~S. Dhillon, Dimitris~S. Papailiopoulos, and Vivienne Sze,
  editors, {\em Proceedings of Machine Learning and Systems 2020, MLSys 2020,
  Austin, TX, USA, March 2-4, 2020}. mlsys.org, 2020.

\bibitem{li2021fedbn}
Xiaoxiao Li, Meirui Jiang, Xiaofei Zhang, Michael Kamp, and Qi Dou.
\newblock Fedbn: Federated learning on non-iid features via local batch
  normalization, 2021.

\bibitem{li2021fedrs}
Xin-Chun Li and De-Chuan Zhan.
\newblock Fedrs: Federated learning with restricted softmax for label
  distribution non-iid data.
\newblock In {\em Proceedings of the 27th ACM SIGKDD Conference on Knowledge
  Discovery \& Data Mining}, pages 995--1005, 2021.

\bibitem{li2021anti}
Yige Li, Xixiang Lyu, Nodens Koren, Lingjuan Lyu, Bo Li, and Xingjun Ma.
\newblock Anti-backdoor learning: Training clean models on poisoned data.
\newblock {\em Advances in Neural Information Processing Systems}, 34, 2021.

\bibitem{DBLP:conf/nips/LinKSJ20}
Tao Lin, Lingjing Kong, Sebastian~U. Stich, and Martin Jaggi.
\newblock Ensemble distillation for robust model fusion in federated learning.
\newblock In Hugo Larochelle, Marc'Aurelio Ranzato, Raia Hadsell,
  Maria{-}Florina Balcan, and Hsuan{-}Tien Lin, editors, {\em Advances in
  Neural Information Processing Systems 33: Annual Conference on Neural
  Information Processing Systems 2020, NeurIPS 2020, December 6-12, 2020,
  virtual}, 2020.

\bibitem{liu2019large}
Ziwei Liu, Zhongqi Miao, Xiaohang Zhan, Jiayun Wang, Boqing Gong, and Stella~X
  Yu.
\newblock Large-scale long-tailed recognition in an open world.
\newblock In {\em Proceedings of the IEEE/CVF Conference on Computer Vision and
  Pattern Recognition}, pages 2537--2546, 2019.

\bibitem{mcmahan2017communication}
Brendan McMahan, Eider Moore, Daniel Ramage, Seth Hampson, and Blaise~Aguera y
  Arcas.
\newblock Communication-efficient learning of deep networks from decentralized
  data.
\newblock In {\em Artificial Intelligence and Statistics}, pages 1273--1282.
  PMLR, 2017.

\bibitem{menon2021longtail}
Aditya~Krishna Menon, Sadeep Jayasumana, Ankit~Singh Rawat, Himanshu Jain,
  Andreas Veit, and Sanjiv Kumar.
\newblock Long-tail learning via logit adjustment.
\newblock In {\em International Conference on Learning Representations}, 2021.

\bibitem{DBLP:conf/iclr/ReddiCZGRKKM21}
Sashank~J. Reddi, Zachary Charles, Manzil Zaheer, Zachary Garrett, Keith Rush,
  Jakub Kone{\v{c}}n{\'y}, Sanjiv Kumar, and Hugh~Brendan McMahan.
\newblock Adaptive federated optimization.
\newblock In {\em 9th International Conference on Learning Representations,
  {ICLR} 2021, Virtual Event, Austria, May 3-7, 2021}. OpenReview.net, 2021.

\bibitem{DBLP:journals/corr/abs-2006-16765}
Tao Shen, Jie Zhang, Xinkang Jia, Fengda Zhang, Gang Huang, Pan Zhou, Fei Wu,
  and Chao Wu.
\newblock Federated mutual learning.
\newblock {\em CoRR}, abs/2006.16765, 2020.

\bibitem{DBLP:journals/corr/abs-2104-14729}
Lv Tang.
\newblock Cosformer: Detecting co-salient object with transformers.
\newblock {\em CoRR}, abs/2104.14729, 2021.

\bibitem{DBLP:conf/accv/Tang020}
Lv Tang and Bo Li.
\newblock {CLASS:} cross-level attention and supervision for salient objects
  detection.
\newblock In Hiroshi Ishikawa, Cheng{-}Lin Liu, Tom{\'{a}}s Pajdla, and Jianbo
  Shi, editors, {\em Computer Vision - {ACCV} 2020 - 15th Asian Conference on
  Computer Vision, Kyoto, Japan, November 30 - December 4, 2020, Revised
  Selected Papers, Part {III}}, volume 12624 of {\em Lecture Notes in Computer
  Science}, pages 420--436. Springer, 2020.

\bibitem{tang2022re}
Lv Tang, Bo Li, Senyun Kuang, Mofei Song, and Shouhong Ding.
\newblock Re-thinking the relations in co-saliency detection.
\newblock {\em IEEE Transactions on Circuits and Systems for Video Technology},
  2022.

\bibitem{DBLP:conf/iccv/TangLZDS21}
Lv Tang, Bo Li, Yijie Zhong, Shouhong Ding, and Mofei Song.
\newblock Disentangled high quality salient object detection.
\newblock In {\em 2021 {IEEE/CVF} International Conference on Computer Vision,
  {ICCV} 2021, Montreal, QC, Canada, October 10-17, 2021}, pages 3560--3570.
  {IEEE}, 2021.

\bibitem{van2008visualizing}
Laurens Van~der Maaten and Geoffrey Hinton.
\newblock Visualizing data using t-sne.
\newblock {\em Journal of machine learning research}, 9(11), 2008.

\bibitem{DBLP:conf/iclr/WangYSPK20}
Hongyi Wang, Mikhail Yurochkin, Yuekai Sun, Dimitris~S. Papailiopoulos, and
  Yasaman Khazaeni.
\newblock Federated learning with matched averaging.
\newblock In {\em 8th International Conference on Learning Representations,
  {ICLR} 2020, Addis Ababa, Ethiopia, April 26-30, 2020}. OpenReview.net, 2020.

\bibitem{wang2021field}
Jianyu Wang, Zachary Charles, Zheng Xu, Gauri Joshi, H.~Brendan McMahan,
  Blaise~Aguera y Arcas, Maruan Al-Shedivat, Galen Andrew, Salman Avestimehr,
  Katharine Daly, Deepesh Data, Suhas Diggavi, Hubert Eichner, Advait Gadhikar,
  Zachary Garrett, Antonious~M. Girgis, Filip Hanzely, Andrew Hard, Chaoyang
  He, Samuel Horvath, Zhouyuan Huo, Alex Ingerman, Martin Jaggi, Tara Javidi,
  Peter Kairouz, Satyen Kale, Sai~Praneeth Karimireddy, Jakub Konecny, Sanmi
  Koyejo, Tian Li, Luyang Liu, Mehryar Mohri, Hang Qi, Sashank~J. Reddi, Peter
  Richtarik, Karan Singhal, Virginia Smith, Mahdi Soltanolkotabi, Weikang Song,
  Ananda~Theertha Suresh, Sebastian~U. Stich, Ameet Talwalkar, Hongyi Wang,
  Blake Woodworth, Shanshan Wu, Felix~X. Yu, Honglin Yuan, Manzil Zaheer, Mi
  Zhang, Tong Zhang, Chunxiang Zheng, Chen Zhu, and Wennan Zhu.
\newblock A field guide to federated optimization, 2021.

\bibitem{DBLP:conf/nips/WangLLJP20}
Jianyu Wang, Qinghua Liu, Hao Liang, Gauri Joshi, and H.~Vincent Poor.
\newblock Tackling the objective inconsistency problem in heterogeneous
  federated optimization.
\newblock In Hugo Larochelle, Marc'Aurelio Ranzato, Raia Hadsell,
  Maria{-}Florina Balcan, and Hsuan{-}Tien Lin, editors, {\em Advances in
  Neural Information Processing Systems 33: Annual Conference on Neural
  Information Processing Systems 2020, NeurIPS 2020, December 6-12, 2020,
  virtual}, 2020.

\bibitem{wang2021addressing}
Lixu Wang, Shichao Xu, Xiao Wang, and Qi Zhu.
\newblock Addressing class imbalance in federated learning.
\newblock In {\em Proceedings of the AAAI Conference on Artificial
  Intelligence}, pages 10165--10173, 2021.

\bibitem{yurochkin2019bayesian}
Mikhail Yurochkin, Mayank Agarwal, Soumya Ghosh, Kristjan Greenewald,
  Trong~Nghia Hoang, and Yasaman Khazaeni.
\newblock Bayesian nonparametric federated learning of neural networks, 2019.

\bibitem{zhang2021practical}
Jie Zhang, Chen Chen, Bo Li, Lingjuan Lyu, Shuang Wu, Jianghe Xu, Shouhong
  Ding, and Chao Wu.
\newblock A practical data-free approach to one-shot federated learning with
  heterogeneity.
\newblock {\em arXiv preprint arXiv:2112.12371}, 2021.

\bibitem{Zhang_2022_CVPR}
Jie Zhang, Bo Li, Jianghe Xu, Shuang Wu, Shouhong Ding, Lei Zhang, and Chao Wu.
\newblock Towards efficient data free black-box adversarial attack.
\newblock In {\em Proceedings of the IEEE/CVF Conference on Computer Vision and
  Pattern Recognition (CVPR)}, pages 15115--15125, June 2022.

\bibitem{zhang2020fedpd}
Xinwei Zhang, Mingyi Hong, Sairaj Dhople, Wotao Yin, and Yang Liu.
\newblock Fedpd: A federated learning framework with optimal rates and
  adaptivity to non-iid data, 2020.

\bibitem{zhang2021bag}
Yongshun Zhang, Xiu-Shen Wei, Boyan Zhou, and Jianxin Wu.
\newblock Bag of tricks for long-tailed visual recognition with deep
  convolutional neural networks.
\newblock In {\em Proceedings of the AAAI Conference on Artificial
  Intelligence}, pages 3447--3455, 2021.

\bibitem{zhao2018federated}
Yue Zhao, Meng Li, Liangzhen Lai, Naveen Suda, Damon Civin, and Vikas Chandra.
\newblock Federated learning with non-iid data, 2018.

\bibitem{Zhong_2022_CVPR}
Yijie Zhong, Bo Li, Lv Tang, Senyun Kuang, Shuang Wu, and Shouhong Ding.
\newblock Detecting camouflaged object in frequency domain.
\newblock In {\em Proceedings of the IEEE/CVF Conference on Computer Vision and
  Pattern Recognition (CVPR)}, pages 4504--4513, June 2022.

\bibitem{DBLP:journals/corr/abs-2110-12748}
Yijie Zhong, Bo Li, Lv Tang, Hao Tang, and Shouhong Ding.
\newblock Highly efficient natural image matting.
\newblock {\em CoRR}, abs/2110.12748, 2021.

\end{thebibliography}
}

\end{document}